\documentclass[letterpaper, 10 pt, conference]{ieeeconf}  

\IEEEoverridecommandlockouts                              

\usepackage{graphics} 
\usepackage{amsmath} 
\usepackage{amssymb}  

\usepackage{graphicx}

 \newcommand{\cmark}{Y}%
 \newcommand{\xmark}{N}%

\title{\LARGE \bf
Pseudo-Labeling and Contextual Curriculum Learning for Online Grasp Learning in Robotic Bin Picking
}
\author{Huy Le$^{1,2}$, 
        Philipp Schillinger$^{2}$,
		Miroslav Gabriel$^{2}$,
		Alexander Qualmann$^{2}$,
		Ngo Anh Vien$^{2}$
\thanks{$^{1}$ Technical University Dortmund }
\thanks{$^{2}$ Bosch Center for Artificial Intelligence (BCAI), Germany. Email: firstname.lastname@de.bosch.com}
}
\begin{document}
\maketitle
\thispagestyle{empty}
\pagestyle{empty}

  \renewcommand{\baselinestretch}{.964}
   \setlength{\abovedisplayskip}{0pt}
   \setlength{\belowdisplayskip}{0pt}
   \setlength{\belowcaptionskip}{0pt}

\maketitle

\begin{abstract}
The prevailing grasp prediction methods predominantly rely on offline learning, overlooking the dynamic grasp learning that occurs during real-time adaptation to novel picking scenarios. These scenarios may involve previously unseen objects, variations in camera perspectives, and bin configurations, among other factors. In this paper, we introduce a novel approach, SSL-ConvSAC, that combines semi-supervised learning and reinforcement learning for online grasp learning. By treating pixels with reward feedback as labeled data and others as unlabeled, it efficiently exploits unlabeled data to enhance learning. In addition, we address the imbalance between labeled and unlabeled data by proposing a contextual curriculum-based method. We ablate the proposed approach on real-world evaluation data and demonstrate promise for improving online grasp learning on bin picking tasks using a physical 7-DoF Franka Emika robot arm with a suction gripper. Video: https://youtu.be/OAro5pg8I9U
\end{abstract}
%

\section{Introduction}
\label{intro}
The core task in robotic manipulation is grasping, a fundamental skill that opens doors to more complex actions like pick and place or bin picking \cite{1309.2660}. In bin picking, the goal is to take objects out of a container and put them in specific places, which has wide applications. However, bin picking is challenging due to issues like noisy perception, object obstructions, and collisions in planning. Thus, there is a need for a robust approach to handle this task effectively \cite{kleeberger2020survey}. To address these challenges, modern grasping techniques have harnessed advanced deep learning methods. These techniques empower the model to predict grasping actions without relying on predefined models, thereby making them applicable to a broad spectrum of objects in unstructured environments  \cite{kleeberger2020survey,newbury2023deep}. However, it is worth noting that most of the approaches discussed in the literature depend on supervised learning and offline training, potentially limiting their ability to adapt to unseen objects or new environmental conditions \cite{danielczuk2020exploratory,fu2022legs}.

To address these challenges, this paper's primary emphasis lies in addressing the issue of online grasp learning, which is often framed as a reinforcement learning problem \cite{1803.09956,BerscheidMK19,BerscheidFK21,2111.01510}. These works have commonly used a fully
convolutional network (FCN) to learn dense pixel-wise grasp quality predictions, i.e. the critic. The pixel-wise parameterization is also used for the grasp primitive, i.e. the actor \cite{abs-2305-03942}. However during online learning, the agent receives sparse feedback of grasp success or failure at only one pixel location selected by the policy. The networks get updated accordingly through backpropagation via the loss at this only pixel point. 

\begin{figure}[ht]
    \centering
    \includegraphics[width=0.8\columnwidth]{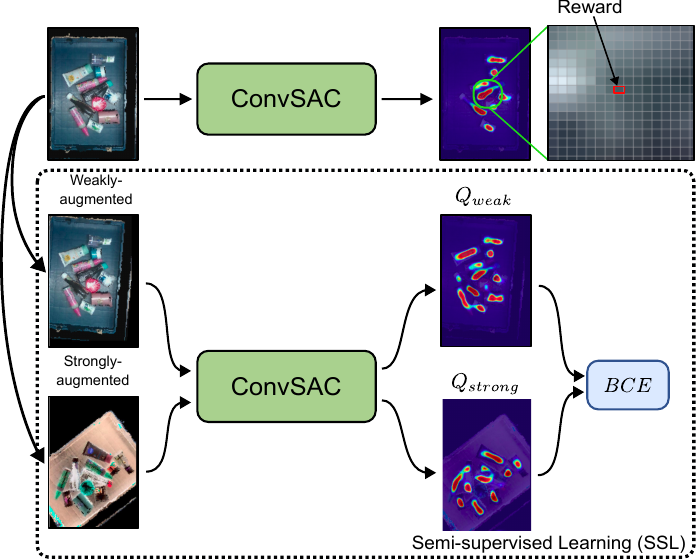}
    \caption{SSL-ConvSAC: A combined SSL and ConvSAC grasp learning approach to address the sparse reward feedback problem in online grasp learning. During online learning, only one pixel point gets feedback, hence the loss is sparsely backpropagated. In contrast, SSL-ConvSAC will learn from both ground-truth and pseudo-labeled reward feedback.}
    \label{fig:overview_pipeline}
\end{figure}

In this paper, we propose an approach that is able to take advantages of backpropagation via the whole pixel points at each training step. In particular, we combine the advantages of semi-supervised learning (SSL) and RL-based online grasp learning. We cast the pixel point with reward feedback as labeled data, while the remaining pixels without reward feedback as unlabeled data. SSL allows us to enable for exploitation of unlabeled data to improve training progress and overall performance \cite{sohn2020fixmatch,zhang2021flexmatch}. To this end, we propose SSL-based fully Convolutional Soft-Actor Critic (SSL-ConvSAC), that combines both true rewards and pseudo-labeled rewards for grasp policy learning as depicted in Fig.~\ref{fig:overview_pipeline}.
	
To summarize, our main contributions are as follows:
    \begin{itemize}
    \item We address an unexplored problem in online grasp learning, \emph{sparse reward feedback}. We propose a principled approach SSL-ConvSAC, that take advantages of unlabeled data to improve the learning efficiency.    
    \item We show that different SSL methods can be integrated into SSL-ConvSAC. This integration also enables for curriculum learning-based SSL methods.
    \item We propose a contextual curriculum-based SSL-ConvSAC method to mitigate the data imbalance issue. We observe an extreme imbalance issue between the amount of labeled and unlabeled data. This problem could cause the training to diverge. 
    \item We evaluate and ablate the proposed methods on real-world evaluation data and online grasp learning on bin picking tasks using a physical 7-DoF Franka Emika robot arm with a suction gripper.
    \end{itemize}

\section{Related Work}
\label{related_work}

\subsection{Modern Robot Grasping Methods}

Modern robot grasping methods are known for applications of deep learning techniques and training on a large amount of available data. Mahler et al. \cite{Dex} and Zeng et al. \cite{zeng2022robotic} propose to predict grasp maps for both suction and parallel-jaw grasps through the employment of supervised datasets with RGB-D or depth. In a similar context, Morrison et al. \cite{MorrisonLC18} and Satish et al. \cite{satish2019policy} have adopted a comparable approach, focusing on predicting pixel-wise grasp quality maps and 4-DoF parallel-jaw grasp configurations. On the other hand, alternative strategies involve the utilization of point clouds \cite{fang2020graspnet, sundermeyer2021contact} aiming to infer dense pixel-wise grasp qualities and gripper configurations. Recent advancements have also introduced the concept of pixel-wise grasp maps and predictions for grasp configurations tailored to single-suction grippers \cite{cao2021suctionnet} and multi-suction cup grippers \cite{abs-2307-16488}.

\subsection{Deep RL-based Grasping Methods}

Vision-based Reinforcement Learning (RL) has emerged as a promising approach for enhancing robotic grasping. This technique involves harnessing visual information, such as RGB-D or point cloud data, to guide the actions of robots using RL networks \cite{sur}. RL-based methods \cite{1806.10293,levine2018learning} typically adopt an end-to-end approach, wherein they optimize closed-loop policies for grasp planning directly from raw visual inputs. However, a significant challenge lies in the necessity for large quantities of high-quality training data, primarily due to the high dimensionality of visual \cite{review} or depth \cite{8463204} inputs. In addition to closed-loop RL methods, open-loop RL approaches are also applied in scenarios involving 6-DoF bin picking \cite{1803.09956,BerscheidFK21,2111.01510}.
Several studies have explored the development of advanced online grasp learning techniques \cite{fu2022legs,danielczuk2020exploratory,li2020accelerating,laskey2015multi}. However, their applications have primarily been demonstrated in scenarios involving single, isolated objects. One line of research \cite{shi2023uncertaintydriven} focuses on online exploratory grasp learning for the new scenes. Another line of research applying Equivariant Neural Network \cite{zhu2022grasp} archived good grasping performance. 

\subsection{Semi-Supervised Learning Methods}

Consistency regularization and pseudo-labeling are two popular methods from Semi-supervised learning (SSL). FixMatch \cite{sohn2020fixmatch}, a recently introduced novel approach, achieves competitive results by incorporating both weak and strong data augmentations, and the cross-entropy loss as the criterion for consistency regularization. Follow-up works propose improvements by introducing curriculum learning to FixMatch such as FlexMatch \cite{zhang2021flexmatch}, FreeMatch \cite{Wang2022FreeMatchST}, SoftMatch \cite{chen2023softmatch}. An alternative direction to improve FixMatch is to use pseudo labels to train a gentle teaching assistant (GTA) network. The student network only exploits knowledge from the GTA's feature extractor. 

There have been applications of SSL, which are similar to our use-case, for object detection \cite{li2022pseco,li2022dtg,xu2021end}, and for problems with sparsely annotated image data \cite{rambhatla2022sparsely,wang2023calibrated,yoon2021semi,wang2021co}. However we observe that the online grasp learning problem has far more extremely sparse annotation.





\section{Problem Statement}
\label{problem-statement}
This paper considers the online grasp learning problem for bin picking application. Given an RGB-D image of the scene $I \in R^{H\times W\times 4}$, we aim to online learn a grasping policy $\pi$ that is a mapping from $I$ to output map $\in R^{H\times W\times 4}$ that maximizes the long-term total grasp success rate. The policy output is a multi-channel map of pixel-wise 1-dim grasp quality $Q$ and pixel-wise 3-dim grasp configurations $A$ representing gripper rotation via Euler angles, which is commonly used in previous work \cite{zeng2022robotic,abs-2307-16488}. The action having the best grasp quality is selected to execute, specifically the selected pixel location is $(h^{\ast},w^{\ast})=\arg\max_{h',w'}Q[h',w']$, and the grasp configuration is extracted from the action map as $A[h^{\ast},w^{\ast}]$. After each grasp attempt, the reward $r_t$ is 1 if it succeeds in picking an object, otherwise 0. As a result, the policy $\pi(I)$ is optimized to maximize the total grasp success return $\sum_t r_t$. The reward feedback $r$ is given to only the selected grasp, i.e. at pixel $(h^{\ast},w^{\ast})$, where other pixel locations $\{h,w\}_{(h,w)\in H\times W, (h,w) \# (h^{\ast},w^{\ast})}$ do not obtain reward feedback. We frame this problem as {\bf sparse reward feedback} with respect to the nature of extremely unbalanced ratio between the two type of labeled and unlabeled data. To improve data efficiency, our approach directly exploits both data with reward feedback at $(h^{\ast},w^{\ast})$ and without reward feedback at $\{h,w\}_{(h,w)\in H\times W, (h,w) \# (h^{\ast},w^{\ast})}$. The proposed method is based on the synergy between pseudo-labeling for semi-supervised learning and reinforcement learning.

\section{Methodology}
\label{method}
\begin{figure*}[ht]
    \centering
    \includegraphics[width=0.85\linewidth]{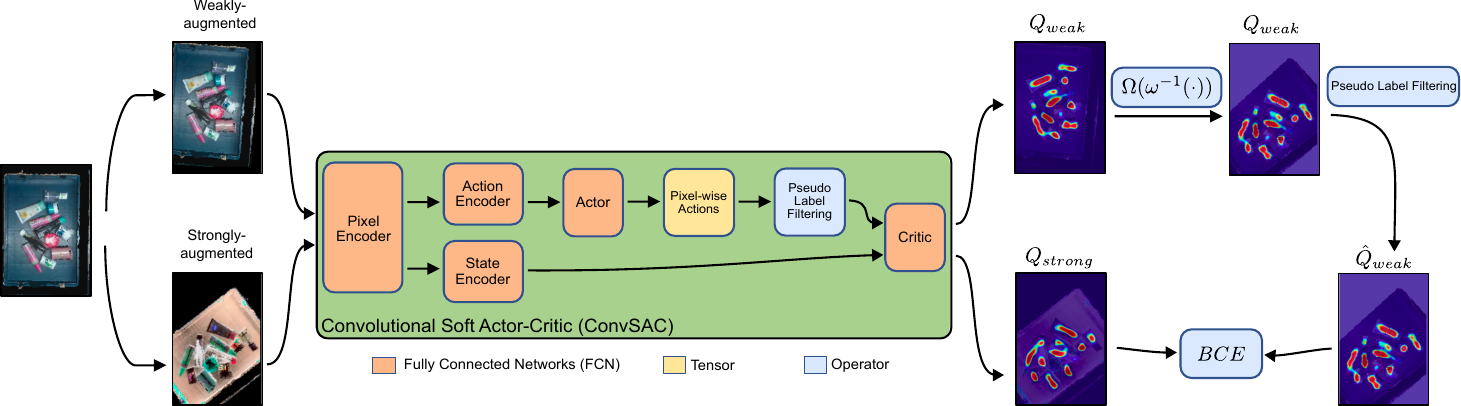}
    \caption{SSL-ConvSAC Pipeline: Both strongly and weakly augmented images are fed to a ConvSAC network. The pixel-wise prediction of a weakly augmented input is used to provide pseudo labels if their confidence are above a threshold. These pixel-wise pseudo-labeled rewards are then used to compute consistency regularization to update both the Critic and Actor receiving the strongly augmented image as input.}
    \label{fig:sample_double}
\end{figure*}

\subsection{Problem Formulation}
\label{formulation}
In this paper, we propose to formulate this online grasp learning problem as a Markov decision process (MDP), ($\mathcal{S}, \mathcal{A}, \mathcal{P}, \mathcal{R}$), where $\mathcal{S}$ is the state space, $\mathcal{A}$ is the action space, $\mathcal{P}$ is the transition probability function, and $\mathcal{R}$ is the reward function. We adopt the fully convolutional network (FCN) which is the same architecture used by ConvSAC \cite{2111.01510} and HACMan \cite{abs-2305-03942} to infer the dense grasp configuration map $A_\phi(s)$ via an actor network and approximate the dense grasp quality map via a critic network $Q_\theta(s, A_\phi(s))$. The network infers an embedding for each pixel location using a {\bf Pixel Encoder} network. The {\bf Actor} module views each pixel as a distinct state and convolves over encoded pixels and infers a Gaussian action for each of them. Actions are concatenated with their corresponding pixel embedding and evaluated by the {\bf Critic} module resulting in a Q-value map $Q_\theta$.

Similar to the setting of ConvSAC \cite{2111.01510}, the state $s$ is represented by a 7-dim input data, which is composed of a color image $I_c$, a normal surface map $I_n$, and a height map $I_c$, $s_t = (I_c, I_n, I_d)_t$ with $I_c \in \mathbb{R}^{H*W*3}$, $I_n \in \mathbb{R}^{H*W*3}$ and $I_d \in \mathbb{R}^{H*W*1}$. In our experiments, the states are captured by a stereo sensor with a top-down view of the object bin. As a RL-based grasp learning approach, we maintain a replay buffer of samples $\{s_t, a_t, r_t\}$, where $a_t=A_t[h_t,w_t]$ is an action at the only selected or \emph{labeled} pixel $(h_t,w_t)$. The critic and actor networks are updated similar to ConvSAC \cite{2111.01510}, where the critic loss is formulated as a classification task with reward labels $r \in \{0,1\}$ denoting grasp failure or success. The critic uses a BCE loss and the episode horizon terminates after each grasp attempt. Specifically, we define the critic and actor losses for the \emph{labeled} pixels as follows:
\begin{equation}
\begin{aligned}
\mathcal{L}^l_{\text {critic}} &= \operatorname{BCE}\left(Q_t\left(s_t, a_t\right), r_t\right) \\
\mathcal{L}^l_{\text {actor}} &= \alpha \operatorname{log} \pi\left(a_{t} | s_{t} \right) - Q_t\left(s_{t}, a_{t}\right),
\end{aligned}
\label{convsac-updates}
\end{equation}
where $\alpha$ is an entropy regularization coefficient. Note that this update backpropagates the loss through only one pixel $(h_t,w_t)$ at both the grasp quality map $Q$ and action map $A$.

We propose to tackle the problem of sparse reward feedback in RL-based online grasp learning through semi-supervised learning. In our particular problem, we are facing an extremely unbalanced data issue. Specifically, for each input image the amount of labeled data is only $N_l = 1$ while the amount of unlabeled data is $N_u = (H \times W) - 1$ is the remaining state pixels. This setting is due to the fact that rearranging the scene to the previous state in order to collect grasp samples at other pixel locations can result in a different state in a real-world setup. In addition, we tackle a realistic  setting where online learning operates on an industrial picking cell without interruption.

\subsection{General Approach}
Our main contribution is to i) first leverage state-of-the-art SSL techniques for the online grasp learning problem, such as FixMatch \cite{sohn2020fixmatch} and curriculum learning-based SSL FlexMatch \cite{zhang2021flexmatch} and FreeMatch \cite{Wang2022FreeMatchST}, and ii) propose a new contextual curriculum learning-based SSL. We introduce a general approach that enables the integration of different SSL methods, called SSL-ConvSAC, as depicted in Fig.~\ref{fig:sample_double}.

We adopt consistency regularization for SSL to rewrite the losses of the actor $A_\phi$ and critic $Q_\theta$.
The critic and actor models are updated using a joint objective based on labeled and unlabeled data. The updates using labeled data are defined in Eq.~\ref{convsac-updates}. The updates using unlabeled data are written as in Eq.~\ref{ssl-convsac-updates}, given a data sample $(s,a,r)$ where action $a$ encodes labeled pixel $(h,w)$ with reward $r$ while unlabeled pixels are $U=\{h',w'\}_{(h',w')\in H\times W, (h',w') \# (h,w)}$. 

\begin{equation}
\begin{aligned}
\mathcal{L}^u_{\text {critic}} &= \frac{1}{N_u}\lambda(\hat{Q};U) \operatorname{BCE} \left( \hat{Q} , Q(\hat{s}, \pi(\hat{s})) \right) \\
\mathcal{L}^u_{\text {actor}} &= \frac{1}{N_u}\lambda(\hat{Q};U)\left( \alpha \operatorname{log} \pi\left(A | \hat{s} \right) - Q\left(\hat{s}, \pi(\hat{s})\right)
\right)
\end{aligned}
\label{ssl-convsac-updates}
\end{equation}
where $\hat{s} = \Omega(s)$ denotes a strongly-augmented data given input $s$. $\hat{Q}= \left[Q(\omega(s),\pi(\omega(s)) > 0.5\right]$ computes pixel-wise pseudo labels, i.e. $\{0,1\}$, where $\omega(s)$ is a weakly-augmented data. $Q(\Omega(s), \pi(\Omega(s)))$ and $\pi(\Omega(s))$ compute a grasp quality map and an action map of the strongly-augmented data. The weight $\lambda(\hat{Q};U) \in R^{H*W}$ is a pixel-wise weighting function that can be defined differently according to a different choice of an SSL method. The conditioning on $U$ means that only unlabeled pixels matter in this operation, i.e. $\lambda(\hat{Q};U)$ has a zero value at labeled pixel $(h,w)$ and values in range $[0,1]$ at unlabeled pixels. Please note that our suggested SSL objective computes the pixel-wise loss (including BCE with no reduction), enabling it to utilize parallel computations in fully convolutional networks to process the loss for all $N_u$ unlabeled data points simultaneously.

As a result, the joint objectives of the actor and critic are $\mathcal{L}_{\text {critic}}=\mathcal{L}^l_{\text {critic}} + \mathcal{L}^u_{\text {critic}}$ and $\mathcal{L}_{\text {actor}}=\mathcal{L}^l_{\text {actor}} + \mathcal{L}^u_{\text {actor}}$, respectively. The SSL objective for unlabeled data is pixel-wise computed, therefore the final loss is then a sum of losses over all pixels. 

Note that whenever there are $\arg\max$ or $\max$ operations on grasp quality map $Q \in R^{H\times W}$, we implicitly assume $Q \in R^{H\times W\times 2}$ for binary classes, specifically, $Q[:,:,1] = Q$ which is $Q$-value for class \emph{success}, $Q[:,:,0] = 1.0 - Q $ for class \emph{failure}. And $\arg\max$ or $\max$ operations are applied across the last axis.
\subsubsection{FixMatch-based SSL-ConvSAC}
We first propose to leverage FixMatch \cite{sohn2020fixmatch} for SSL-ConvSAC. We define a constant threshold $\tau$ based on which pseudo-labels with high confidence will be retained. In particular,
the weighting function is computed as follows
\begin{align}
     \lambda(\hat{Q}_t;U_t) = \mathbb{I}\left( \max(\hat{Q}_t) \ge \tau \right)
     \label{weighting}
\end{align}
where $\mathbb{I}$ is an identity function.

\subsubsection{Curriculum-based SSL-ConvSAC}
We leverage two curriculum-based SSL frameworks FlexMatch \cite{zhang2021flexmatch} and FreeMatch \cite{Wang2022FreeMatchST}. Instead of using a fixed constant threshold $\tau$, FlexMatch and FreeMatch introduce curriculum learning to tune $\tau$ in order to control the way pseudo labels from individual class are retained. The following proposals computes an adaptive threshold that can be used for re-computing the weighting function specifically for each class $c$ as follows.
\begin{align}
     \lambda_t(\hat{Q}_t;U_t) = \mathbb{I}\left( \max(\hat{Q}_t) \ge \tau_t(\arg\max \hat{Q}_t) \right)
 \label{curriculum:lambda}
\end{align}
where $\tau_t$ will be adapted according to curriculum learning.

\paragraph{FlexMatch-based SSL-ConvSAC}
Similar to original FlexMatch, we propose to model learning effect $\sigma_{t}(c)$, $c\in\{0,1\}$ at each training step $t$, where class with fewer samples having their prediction confidence reach the threshold is considered to have a greater learning difficulty or a worse learning status. Assuming that the size of the replay buffer is $|B|$, then the total number of unlabeled pixels is $N_u \times|B|$. The learning effect is computed as follows
\begin{equation}
\begin{aligned}
\sigma_t(c)=\sum_{s \in B} \sum_{n=1}^{N_u} \mathbb{I}\left(\max Q(\omega(s),\pi(\omega(s))>\tau\right) \\
\cdot \mathbb{I}\left(\arg \max Q(\omega(s),\pi(\omega(s))=c\right).
\end{aligned}    
\label{flexmatch:learning_effect}
\end{equation}
The operation inside the identity function is pixel-wise as usual. The summation takes sum over all unlabeled pixels $n$ and across samples in the replay buffer. As a result, the adaptive threshold $\tau_t(c)$ can now be computed by normalizing $\sigma_t(c)$ in the range $[0,1]$ as follows.
\begin{align}
& \beta_t(c)=\frac{\sigma_t(c)}{\max _c \sigma_t}, \quad  \tau_t(c)=\beta_t(c) \cdot \tau,
\label{flexmatch:threshold}
\end{align}
where normalized learning effect $\beta_t(c)$ is equal to 1 for the best-learned class and lower for the hard classes. We also use a similar warm-up
process and a non-linear mapping function from FlexMatch to enable the thresholds to have a non-linear increasing curve in the range from $0$ to $1$.

\paragraph{FreeMatch-based SSL-ConvSAC}
Instead of adjusting the confidence threshold according to only the current step's information as in FlexMatch,
FreeMatch proposes to self-adapt this value according to the model learning progress. In particular, a self-adaptive global threshold is computed as in Eq.~\ref{freematch:global} to track the overall learning status globally across all classes among unlabeled data.

\begin{align}    
    \tau^{\text{global}}_t = \alpha \tau^{\text{global}}_{t-1} + (1-\alpha)\frac{1}{ N_u |B|}\sum_{s \in B} \sum_{n=1}^{N_u} \max q_s ,
\label{freematch:global}
\end{align}
with $t>0$ and $\tau^{\text{global}}_0 = 1/2$ (as the number of classes is 2, i.e. success or failure), where we denote $q_s = Q(\omega(s),\pi(\omega(s))$ and $\alpha \in (0,1)$ is the momentum decay of the exponential moving average of the confidence. A self-adaptive local threshold to adjust the global threshold in a class-specific fashion is computed as follows.
\begin{align}
    \tilde{p}_t(c) = \alpha \tilde{p}_{t-1}(c) +(1-\alpha) \frac{1}{ N_u |B|}\sum_{s \in B} \sum_{n=1}^{N_u} q_s(c),
    \label{freematch:pt}
\end{align}
with $t>0$ and $\tilde{p}_0 = 1/2$. As a result, the final adaptive threshold for each individual class is computed as
\begin{align}
    \tau_t(c) = \frac{\tilde{p}_t(c)}{\max_{c\in\{\text{failure}, \text{success}\}}\{\tilde{p}_t(c)\}} \cdot\tau^{\text{global}}_t
    \label{freematch:all2}
\end{align}
The fairness regularization introduced in FreeMatch has been excluded from our experiments because it did not demonstrate any advantages.

\subsubsection{Contextual Curriculum-based SSL-ConvSAC}
We observe that the main challenge in our setting compared to standard SSL is at the extreme imbalance between labeled and unlabeled data. This will quickly lead to the confirmation bias problem \cite{arazo2020pseudo}. The authors show that most SSL methods can suffer from this problem if the mini-batch contains a ratio of 1:100 between labeled and unlabeled data. Inspired by \cite{arazo2020pseudo}, we propose two main technical and one fundamental contributions that can improve generalization and reduce confirmation bias. The technical contributions aim to reduce the confidence of the network as suggested by \cite{arazo2020pseudo}. 
\begin{itemize}
    \item {\bf Lower-bounded confidence threshold}: This helps to filter pseudo labels  with low-confidence for curriculum-based methods. In particular, we lower-bound the adaptive threshold as $\tau_t = \max\{\tau_t , \tau_\text{lb}\}$, $\tau_\text{lb}$ is a predefined lower-bound confidence threshold.
    \item {\bf Soft-weighting function}: The current hard-weighting $\lambda_t$ in Eq.~\ref{curriculum:lambda} treats pseudo labels of both low and high confidence equally as long as their confidence is above the threshold. We propose using soft-weighting via a soft-max function: $
    \lambda_t(\hat{Q}_t;U_t) \propto \exp\left( \hat{Q}_t[\mathbb{I}_t] \right)    
    $,
    where $\mathbb{I}_t = \mathbb{I}\left( \max(\hat{Q}_t) \ge \tau_t(\arg\max \hat{Q}_t)\right)$.
    \item {\bf Contextual curriculum-based learning}: All previous SSL-ConvSAC variants compute thresholds adaptively to each class, i.e. in FlexMatch and FreeMatch-based SSL-ConvSAC and these values $\sigma_t, \beta_t, \tau_t, \tilde{p}_t$ are 2-dim. However, we observe that different pixel locations in an input image though having the same class, e.g. success, their grasp quality values are not necessarily identical. 
    We propose to represent $\sigma_t, \beta_t, \tau_t, \tilde{p}_t \in R^{H\times W\times 2}$ to depend on pixel contexts. As a result, we can rewrite Eqs.~\ref{flexmatch:learning_effect}, \ref{freematch:global}, \ref{freematch:pt} as pixel-wise versions as follows,
    \begin{equation}
\begin{aligned}
\sigma_t(c) &=\sum_{s \in B} \mathbb{I}\left(\max Q(\omega(s),\pi(\omega(s))>\tau\right) \\
 &\cdot \mathbb{I}\left(\arg \max Q(\omega(s),\pi(\omega(s))=c\right). \\
 \tau^{\text{global}}_t &= \alpha \tau^{\text{global}}_{t-1} + (1-\alpha)\frac{1}{ |B|}\sum_{s \in B} \max q_s \\
\tilde{p}_t(c) &= \alpha \tilde{p}_{t-1}(c) +(1-\alpha) \frac{1}{ |B|}\sum_{s \in B} q_s(c),
\end{aligned}    
\end{equation}
The final equations Eq.~\ref{flexmatch:threshold}, \ref{freematch:all2} are now pixel-wise computed, too. As a result, the weighting function in Eq.~\ref{curriculum:lambda} involves fully pixel-wise terms.
   
\end{itemize}

\section{Experiment Setting}
\label{experiment}
We carry out two set of experiments: i) a large scale evaluation task using pre-collected data with a setup described in Section \ref{evaluation}; ii) online learning directly on a physical robot system with a setup described in Section \ref{robot_setup}.

\subsection{Weak-Strong Augmentation Pipelines}

Our set of weak and strong augmentation are: 
i) {\bf Color transformation} in RandAugment \cite{DBLP:journals/corr/abs-1909-13719}, which includes AutoContrast, Brightness, Contrast, Equalize, Posterize, Sharpness, and Solarize.
ii) {\bf Geometric transformation}: Rotation and shift operations that alter the information about normal vectors in the seven-channel image, and matching operation \cite{abs-2005-04757}. 
iii) {\bf Noise Operator}: Uniform noise and Binary noise.

We apply color transformation on RGB channels, Uniform noise on Depth channel, Binary noise on normal vectors channels, Geometric transformation on the whole seven channels and grasp action configurations. In the weakly augmentation pipeline, we set random Rotation in $[-10, 10]$ degree, and random shifting with $[-10, 10]$ pixels, color Jittering on RGB channel. Strong augmentation pipeline sets random rotation in the range $[-180, 180]$ degree, and shifting $[-30, 30]$ pixels on the whole seven channels. For the depth channel, we use uniform noise of 5mm range and 10\% zero out in normal vectors.

\subsection{Technical Details}
\subsubsection{Training Setting} We use the same ConvSAC network parameters as in \cite{2111.01510}, and slightly different hyperparameters in FixMatch \cite{sohn2020fixmatch}. Concretely, the optimizer for all experiments is Adam with $(\beta_{1}, \beta_{2}) = (0.9, 0.999)$, weight decay $0.0001$, and learning rate $0.0001$. The momentum decay $\alpha$ in FreeMatch is set to 0.95. We perform an exponential moving average model with momentum of 0.99.

\subsubsection{Offline training}
\begin{figure*}
    \centering
    \includegraphics[width=.31\linewidth]{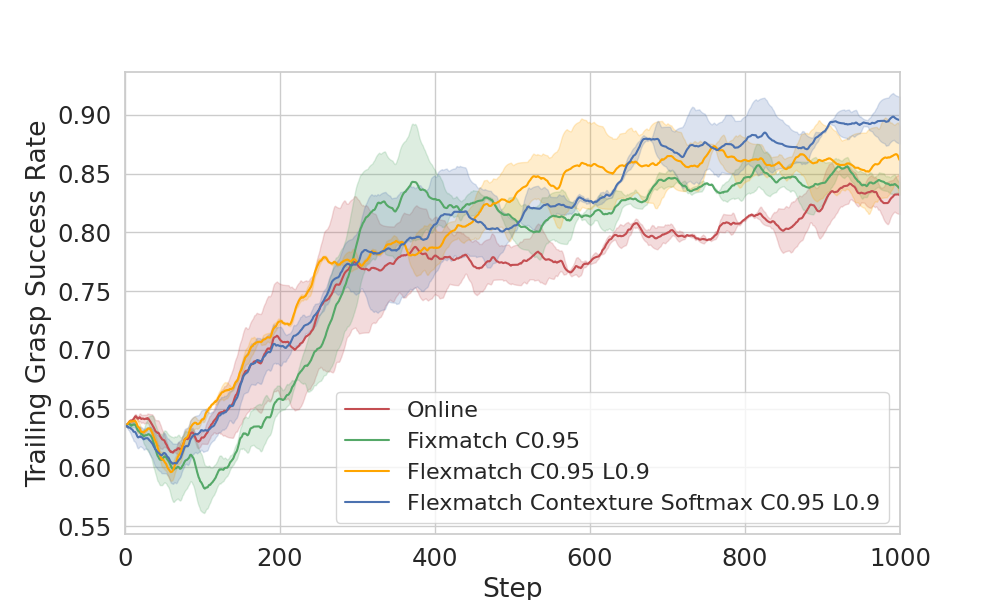}
     \includegraphics[width=0.31\linewidth]{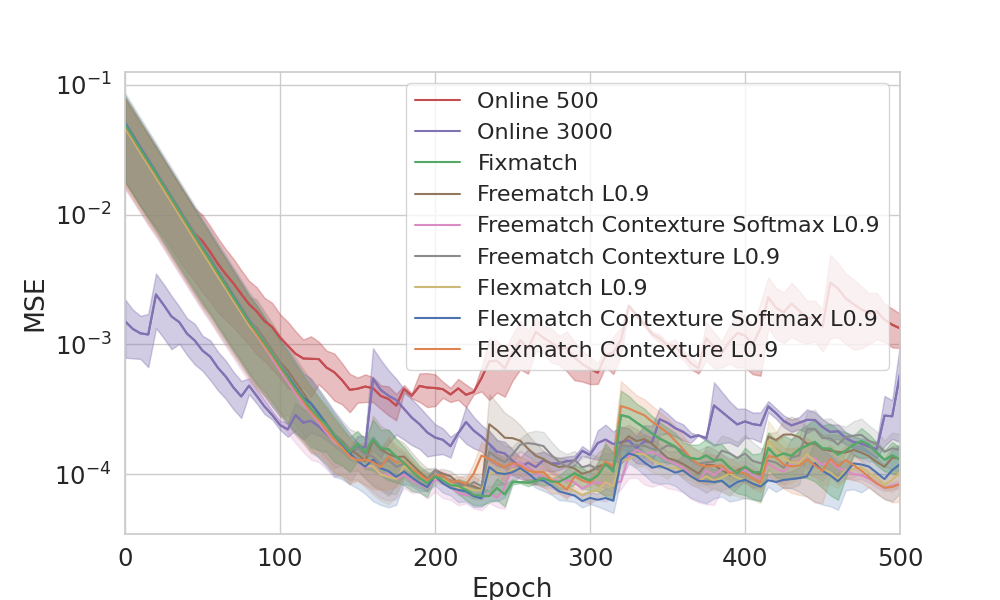} 
     \includegraphics[width=0.31\linewidth]{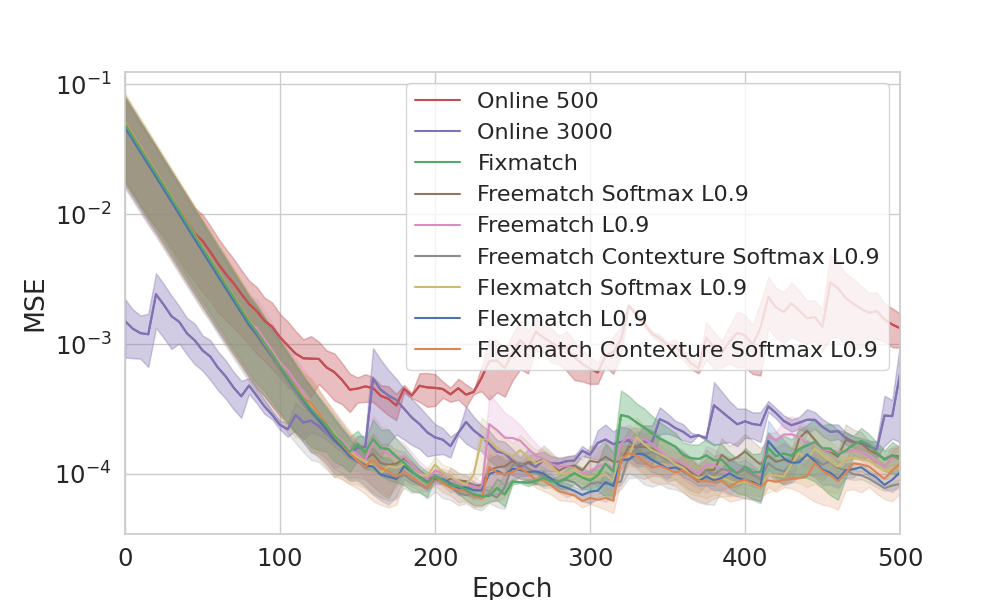}
    \caption{(Left) Online training plots on physical robot. Trailing grasp success rate over the latest 15 bins vs. number of online sample steps. The asynchronous training has a ratio of 10:1 steps per grasp attempt. (Center) Comparisons of contextural models. (Right) Comparisons of soft-weighting models.}
    \label{fig:realrobot_plot}
\end{figure*}

We first train a ConvSAC model offline for 100 epochs to use i) for data collection in the large-scale evaluation task, ii) as a warm-start policy network for online learning on real robots. The offline dataset consists of 72 simple scenes of random 4 objects from the set in Fig. ~\ref{fig:robot_setup} (right). It can be represented by ${\{I, A, Q\}}$, where input image $I$ are defined in Section \ref{formulation}, and $Q$ is the pixel-wise approximated ground truth $Q$-reward map as discussed in \cite{abs-2307-16488}. The action map $A$ pertaining to a suction gripper is assigned to negative surface normal vectors. This pretrained policy is 64\% grasp success as shown in Fig.~\ref{fig:realrobot_plot} (Left) before any online learning starts.

\subsubsection{Large-Scale Evaluation Setup} 
\label{evaluation}
\paragraph{Evaluation data}
 We collected in total 900 online data points using the pre-trained ConvSAC model above. The scene is prepared with randomly 10-12 objects from the online training objects.
 Each data point is a sample with sparse reward feedback $\{s,a,r\}$. We use a replay buffer of 500 points for the training set. We train all models in the evaluation task for 500 epochs on a \emph{Nvidia A100} GPU with a batch size of 4. The evaluation set consists of the remaining 400 data points. Only for this set, for each image $s$ we also generate negative samples from background areas using background subtraction given an empty bin image. The negative samples are ($s$, a random action, $r=0$). Our evaluation metric is the mean squared error (MSE) between the grasp quality prediction and ground-truth reward 0/1.

\paragraph{Comparing methods}
We benchmark our approach against the following configurations:

\begin{enumerate}
    \item \textbf{Online (ON)}: ConvSAC without SSL. We additionally run ON with 3000 data points for benchmarking.
    \item \textbf{FixMatch (FI)}: FixMatch SSL-ConvSAC with a confidence threshold set at $\tau= 0.95 $.
    \item \textbf{FlexMatch (FL)}: FlexMatch SSL-ConvSAC with different lower-bound confidence threshold $\tau_{\text{lb}}= \{0.5, 0.7, 0.9\}$, w./w.o soft-weighting function, and w./w.o contextual curriculum-based learning.
    \item \textbf{FreeMatch (FR)}: FreeMatch SSL-ConvSAC with the same options as FlexMatch SSL-ConvSAC.
\end{enumerate}

All results are averaged over 3 training trials per setting. We report the mean and its standard deviation.

\subsubsection{Physical Robot Setup}
\label{robot_setup}
\begin{figure}
    \centering
    \includegraphics[width=0.69\columnwidth]{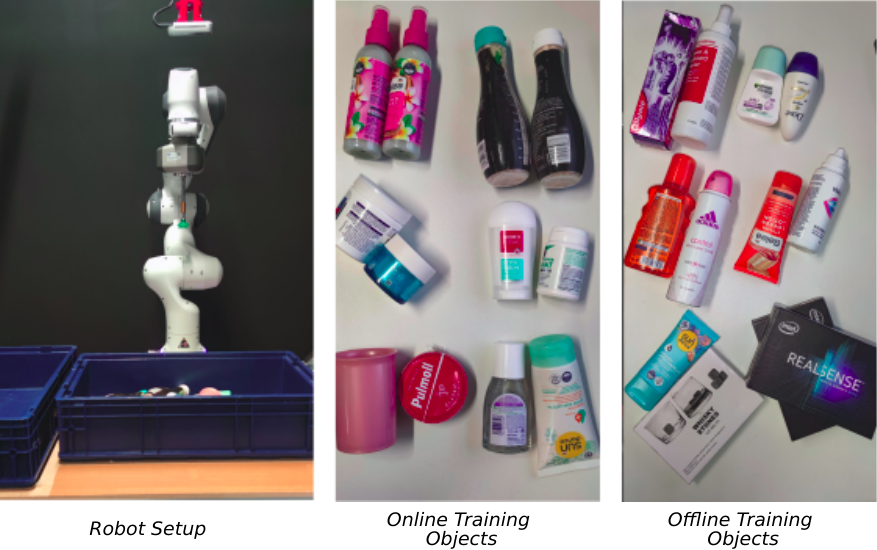}
    \caption{Robot setup (left), online objects (middle), offline objects (right)} 
    \label{fig:robot_setup}
\end{figure}

The experiment was conducted on the \emph{Franka Emika} Robot with \emph{Schmalz} suction gripper and an over the shoulder Realsense d415 camera as seen in Fig.~\ref{fig:robot_setup}. The online training objects consist of 12 common objects depicted in Fig.~\ref{fig:robot_setup} middle. Each online learning scene is prepared with randomly 10-12 objects.

\paragraph{Metrics}
We measure performance based on the \emph{Grasp Success Rate} (SR) and \emph{Bin Completion Rate} (BCR), averaged across the latest 15 bins. Each episode finishes if the bin is clear or the robot has already attempted 15 grasps on this bin. SR is defined as the ratio of successful grasps, and BCR is the ratio of complete bin.

\paragraph{Online training setting} We online learn all comparing models with a learning rate of $(1e-5)$ for both the actor and critic from the pretrained ConvSAC. The model runs on Ubuntu 20.04 and is equipped with an Intel Xeon E3-1505M CPU and a Nvidia GeForce RTX 3090 GPU. 
\paragraph{Comparing methods}
 We run these algorithms:
 i) Online, ii) FlixMatch SSL-ConvSAC with $\tau= 0.95 $, iii) Flexmatch SSL-ConvSAC with contexture, softmax weighting, $\tau_{\text{lb}}=0.9$, and full pseudo-label.  All results are averaged over 2 training trials per setting. We report the mean and its standard deviation.
\begin{table}
\centering
\begin{tabular}{|c|c|c|c|c|c|}
\hline
Methods & S & C & 5 & 100 & Full \\
\hline
ON 500 & & & & &$0.34 \pm 0.13$ \\
ON 3000 & & & &  &$0.11 \pm 0.04$ \\
\hline
FI C$0.95$ & & & & &$0.07 \pm 0.02$ \\
\hline
FR L$0.9$ & \xmark & \xmark & $0.39 \pm 0.13$ & $0.4 \pm 0.2$ & $0.08 \pm 0.02$ \\
FR L$0.9$ & \cmark & \xmark & $0.29 \pm 0.15$ & $0.34 \pm 0.07$ & $0.08 \pm 0.01$ \\
FR L$0.9$ & \xmark & \cmark & $0.38 \pm 0.12$ & $0.32 \pm 0.02$ & $0.08 \pm 0.01$ \\
FR L$0.9$ & \cmark & \cmark & $0.26 \pm 0.03$ & $0.43 \pm 0.19$ & $0.07 \pm 0.02$ \\
FR L$0.7$ & \xmark & \xmark & $0.3 \pm 0.18$ & $0.36 \pm 0.14$ & $0.11 \pm 0.05$ \\
FR L$0.7$ & \cmark & \xmark & $0.36 \pm 0.15$ & $0.57 \pm 0.16$ & $0.1 \pm 0.05$ \\
FR L$0.7$ & \xmark & \cmark & $0.42 \pm 0.03$ & $0.38 \pm 0.1$ & $0.09 \pm 0.02$ \\
FR L$0.7$ & \cmark & \cmark & $0.45 \pm 0.07$ & $0.46 \pm 0.08$ & $0.08 \pm 0.02$ \\
FR L$0.5$ & \xmark & \xmark & $0.35 \pm 0.08$ & $0.41 \pm 0.14$ & $191.81 \pm 148.86$ \\
FR L$0.5$ & \cmark & \xmark & $0.37 \pm 0.2$ & $0.58 \pm 0.16$ & $12.59 \pm 11.89$ \\
FR L$0.5$ & \xmark & \cmark & $0.39 \pm 0.15$ & $0.43 \pm 0.15$ & $191.78 \pm 148.92$ \\
FR L$0.5$ & \cmark & \cmark & $0.32 \pm 0.13$ & $0.38 \pm 0.06$ & $192.51 \pm 162.96$ \\
\hline
FL L$0.9$ & \xmark & \xmark & $0.48 \pm 0.3$ & $0.29 \pm 0.07$ & $0.07 \pm 0.01$ \\
FL L$0.9$ & \cmark & \xmark & $0.41 \pm 0.11$ & $0.47 \pm 0.22$ & $0.08 \pm 0.02$ \\
FL L$0.9$ & \xmark & \cmark & $0.33 \pm 0.24$ & $0.28 \pm 0.07$ & $0.08 \pm 0.02$ \\
FL L$0.9$ & \cmark & \cmark & $0.27 \pm 0.19$ & $0.47 \pm 0.31$ & \textbf{0.06} $\pm$ \textbf{0.01} \\
FL L$0.7$ & \xmark & \xmark & $0.29 \pm 0.1$ & $0.53 \pm 0.28$ & $0.11 \pm 0.05$ \\
FL L$0.7$ & \cmark & \xmark & $0.33 \pm 0.08$ & $0.63 \pm 0.26$ & $0.11 \pm 0.05$ \\
FL L$0.7$ & \xmark & \cmark & $0.46 \pm 0.24$ & $0.46 \pm 0.23$ & $0.07 \pm 0.02$ \\
FL L$0.7$ & \cmark & \cmark & $0.32 \pm 0.08$ & $0.44 \pm 0.12$ & \textbf{0.06} $\pm$ \textbf{0.02} \\
FL L$0.5$ & \xmark & \xmark & $0.35 \pm 0.04$ & $0.38 \pm 0.07$ & $191.76 \pm 149.0$ \\
FL L$0.5$ & \cmark & \xmark & $0.2 \pm 0.03$ & $0.46 \pm 0.24$ & $38.84 \pm 33.39$ \\
FL L$0.5$ & \xmark & \cmark & $0.42 \pm 0.2$ & $0.44 \pm 0.26$ & $191.7 \pm 149.11$ \\
FL L$0.5$ & \cmark & \cmark & $0.38 \pm 0.07$ & $0.57 \pm 0.31$ & $192.46 \pm 162.95$ \\
\hline
\end{tabular}
\caption{MSE on evaluation set ($\times 10^{-3}$): Column S is short for Soft-weighting function, Column C is short for Contexture, and 5, 100, FULL indicate top $k$ pseudo-labels selection. Prefix L indicates $\tau_{\text{lb}}$ value. (Y/N are short for Yes/No)}
\label{tab:mse_eval}
\end{table}
\section{Experiment Results}
\subsection{Large-Scale Evaluation}
\paragraph{Extreme imbalanced data} We first show how extreme the imbalanced data issue is. For each curriculum-based variants, we allows a ratio of 5:1, 10:1 (though many of them with sufficient confidence), and all of pseudo labels accepted to training. The results in Table \ref{tab:mse_eval} show that if the ratio is as low as most settings in SSL literature, the training does not have issues, and performs only comparably as the baseline ON. With an extreme ratio, i.e. \emph{Full}, vanilla curriculum-based methods can diverge if measures from our contextual curriculum-based SSL-ConvSAC are not applied, i.e. $\tau_{\text{lb}}=0.5$ means no lower-bound threshold as the boundary between two classes is at confidence 0.5. The reason is they enroll too many erroneous pseudo-labels with low confidence at the beginning of the training.
\paragraph{Contextual curriculum-based SSL-ConvSAC}
The settings with higher lower bound of 0.7 and 0.9 have consistently achieved good evaluation losses, especially on Full. This shows that this technical fix mitigates the divergence issue and also can take advantages of having more unlabeled data. Finally, the best performing setting is when all measures of contextual curriculum-based SSL-ConvSAC are activated, together with FixMatch SSL-ConvSAC. Fig.~\ref{fig:realrobot_plot} (Center, Right) also shows the training plots of SSL-ConvSACs and baselines. These results tell that SSL methods are able to help training faster and converge to a lower loss compared to baselines Online 500 \& 3000. Again, Contexture \& softmax weighting SSL-ConvSAC achieved the best training progress and the final loss in all settings.

\subsection{Online Learning on Physical Robots}
In physical robot experiments, ours reached an 80\% success rate after just 400 steps, whereas the online model required approximately 600 steps, as illustrated in Fig~\ref{fig:realrobot_plot} (Left). Flexmatch SSL-ConvSAC with contexture, softmax weighting, $\tau_{\text{lb}}=0.9$, and full pseudo-label achieved the best grasp success rate and stable around 90\%. This observation is further supported by Table~\ref{tab:realrobot_bcr}, where the Flexmatch SSL-ConvSAC outperformed Online model with 93.3 \% BCR, and 90\% SR. Note that all training plots have a slight drop at an early stage due to the distribution shift from the pretrained model, i.e. different object portfolio and the offline training scenes have 4 objects instead of 10-12 objects in clutter of online scenes. The out-performance of Flexmatch SSL-ConvSAC compared to non-curriculum FixMatch SSL-ConvSAC, because this online learning setting is based on a pretrained model. This allows the curriculum to benefit from a high confidence model at the beginning, which might not be as high as $\tau=0.95$ used by FixMatch. 
\begin{table}
    \centering
    \begin{tabular}{|c|c|c|c|c|}
        \hline
        Method & Softmax & Contexture & BCR (\%) & SR (\%) \\
        \hline
        ON & & & 60 & 82.5\\
        \hline
        FI C0.95 & & & 70 & 83\\
        \hline
        FL C0.95 L0.9 & \xmark & \xmark & 80 &  86\\
        \hline
        FL C0.95 L0.9 & \cmark & \cmark & \textbf{93.3} & \textbf{90} \\
        \hline
    \end{tabular}
    \caption{Results on physical robots. (Y/N are short for Yes/No)}
    \label{tab:realrobot_bcr}
\end{table}

\section{Conclusion}

This paper introduces SSL-ConvSAC, a novel approach that combines advantages of SSL and RL for online grasp learning. We pose the sparse reward feedback problem for online grasp learning. We show that naively integrating SSL with RL is not enough to tackle this problem, as the amount of unlabeled data in this setting is overwhelming. To mitigate this issue, we propose a contextual curriculum learning approach. We show that the proposed approach is able to fully exploit unlabeled data in our application to improve the overall performance. We demonstrate it on a real-world bin picking setup with a grasp success rate of 90 \%, and bin completion of 93\%. As future work, the topics of pseudo-labeling for closed-loop grasping and online learning on flexible objects would be challenging but bring many interesting applications.

\clearpage
\bibliographystyle{IEEEtran}
\bibliography{my,grasping}

\end{document}